\documentclass[10pt,twocolumn,letterpaper]{article}

\usepackage{cvpr}
\usepackage{times}
\usepackage{epsfig}
\usepackage{graphicx}
\usepackage{amsmath}
\usepackage{amssymb}

\makeatletter
\def\@fnsymbol#1{\ensuremath{\ifcase#1\or \dagger\or \ddagger\or
		\mathsection\or \mathparagraph\or \|\or **\or \dagger\dagger
		\or \ddagger\ddagger \else\@ctrerr\fi}}
\makeatother

\usepackage[pagebackref=true,breaklinks=true,letterpaper=true,colorlinks,bookmarks=false]{hyperref}

\cvprfinalcopy 

\def\cvprPaperID{2703} 

\ifcvprfinal\fi
\begin{document}

\title{Graph-Structured Referring Expression Reasoning in The Wild}

\author{Sibei Yang$^{1}$ \quad\quad Guanbin Li$^{2\dagger}$ \quad\quad Yizhou Yu$^{1,3}$\thanks{Corresponding authors. This work was partially supported by the Hong Kong PhD Fellowship, the Guangdong Basic and Applied Basic Research Foundation under Grant No.2020B1515020048, the National Natural Science Foundation of China under Grant No.61976250 and No.U1811463.}\vspace{2mm}\\
	$^1$The University of Hong Kong  \quad\quad $^2$Sun Yat-sen University \quad\quad $^3$Deepwise AI Lab\\
	{\tt\footnotesize sbyang9@hku.hk}, {\tt\small liguanbin@mail.sysu.edu.cn}, {\tt\small yizhouy@acm.org}
	\vspace{-0mm}
}

\maketitle

\begin{abstract}
Grounding referring expressions aims to locate in an image an object referred to by a natural language expression. The linguistic structure of a referring expression provides a layout of reasoning over the visual contents, and it is often crucial to align and jointly understand the image and the referring expression. In this paper, we propose a scene graph guided modular network (SGMN), which performs reasoning over a semantic graph and a scene graph with neural modules under the guidance of the linguistic structure of the expression. In particular, we model the image as a structured semantic graph, and parse the expression into a language scene graph. The language scene graph not only decodes the linguistic structure of the expression, but also has a consistent representation with the image semantic graph. In addition to exploring structured solutions to grounding referring expressions, we also propose Ref-Reasoning, a large-scale real-world dataset for structured referring expression reasoning. We automatically generate referring expressions over the scene graphs of images using diverse expression templates and functional programs. This dataset is equipped with real-world visual contents as well as semantically rich expressions with different reasoning layouts. Experimental results show that our SGMN\footnote{Data and code are available at \href{https://github.com/sibeiyang/sgmn}{https://github.com/sibeiyang/sgmn}} not only significantly outperforms existing state-of-the-art algorithms on the new Ref-Reasoning dataset, but also surpasses state-of-the-art structured methods on commonly used benchmark datasets. It can also provide interpretable visual evidences of reasoning.
\end{abstract}

\begin{figure}[t]
	\begin{center} 
		\includegraphics[width=1\linewidth]{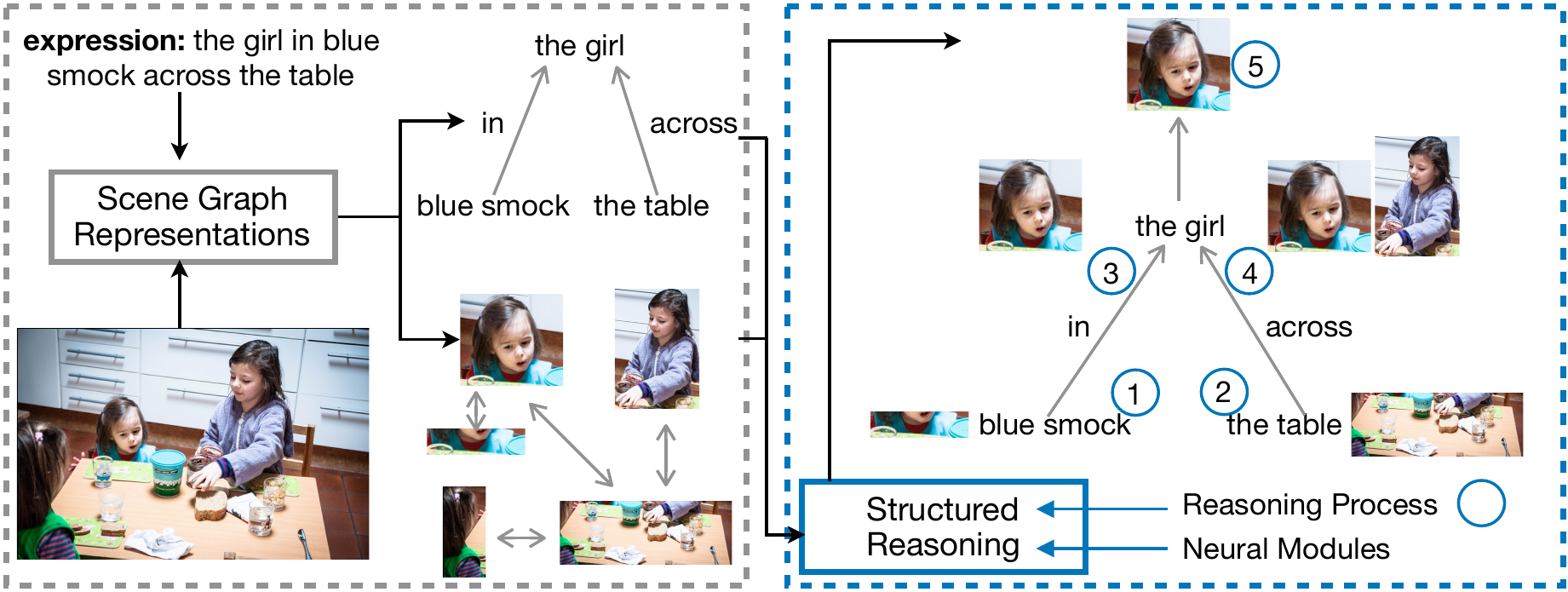}
	\end{center}
	\caption{Scene Graph guided Modular Network (SGMN) for grounding referring expressions. SGMN first parses the expression into a language scene graph and models the image as a semantic graph, then it performs structured reasoning with neural modules under the guidance of the language scene graph.}
	\label{fig:intro}
	\vspace{-0.3cm}
\end{figure}

\section{Introduction}
Grounding referring expressions aims to locate in an image an object referred to by a natural language expression, and the object is called the referent.
It is a challenging problem because it requires understanding as well as performing reasoning over semantics-rich referring expressions and diverse visual contents including objects, attributes and relations.

Analyzing the linguistic structure of referring expressions is the key to grounding referring expressions because they naturally provide the layout of reasoning over the visual contents. For the example shown in Figure~\ref{fig:intro}, the composition of the referring expression ``the girl in blue smock across the table'' (\ie, triplets (``the girl'', ``in'', ``blue smock'') and (``the girl'', ``across'' , ``the table'')) reveals a tree-structured layout of finding the blue smock, locating the table and identifying the girl who is ``in'' the blue smock and meanwhile is ``across'' the table. However, nearly all the existing works either neglect linguistic structures and learn holistic matching scores between monolithic representations of referring expressions and visual contents \cite{luo2017comprehension, zhuang2018parallel, yang2019cross} 
or neglect syntactic information and explore limited linguistic structures via self-attention mechanisms \cite{yu2018mattnet, hong2019learning, yang2019dynamic}.

Consequently, in this paper, we propose a Scene Graph guided modular network (SGMN) to fully analyze the linguistic structure of referring expressions and enable reasoning over visual contents using neural modules under the guidance of the parsed linguistic structure. Specifically, SGMN first models the input image with a structured representation, which is a directed graph over the visual objects in the image. The edges of the graph encode the semantic relations among the objects. 
Second, SGMN analyzes the linguistic structure of the expression by parsing it into a language scene graph~\cite{schuster2015generating, liu2019referring} using an external parser, 
including the nodes and edges of which correspond to noun phrases and prepositional/verb phrases respectively.
The language scene graph not only encodes the linguistic structure but is also consistent with the semantic graph representation of the image. Third, SGMN performs reasoning on the image semantic graph under the guidance of the language scene graph by using well-deigned neural modules~\cite{andreas2016neural, shi2019explainable} including AttendNode, AttendRelation, Transfer, Merge and Norm. The reasoning process can be explicitly explained via a graph attention mechanism.

In addition to methods, datasets are also important for making progress on grounding referring expressions, and various real-world datasets have been released~\cite{kazemzadeh2014referitgame, mao2016generation, yu2016modeling}. However, recent work~\cite{cirik2018visual} indicates dataset biases exist and they may be exploited by the methods. And methods accessing the images only achieve marginally higher performance than a random guess. Existing datasets also have other limitations. First, the samples in the datasets have unbalanced levels of difficulty. Many expressions in the datasets directly describe the referents with attributes due to the annotation process. Such an imbalance makes models learn shallow correlations instead of achieving joint image and text understanding, which defeats the original intention of grounding referring expressions. Second, evaluation is only conducted on final predictions but not on the intermediate reasoning process~\cite{liu2019clevr}, which does not encourage the development of interpretable models \cite{yang2019dynamic, liu2019learning}. Thus, a synthetic dataset over simple 3D shapes with attributes is proposed in \cite{liu2019clevr} to address these limitations. However, the visual contents in this synthetic dataset are too simple, which is not conducive to generalizing trained models on the synthetic dataset to real-world scenes.

To address the aforementioned limitations, we build a large-scale real-world dataset, named Ref-Reasoning. 
We generate semantically rich expressions over the scene graphs of images~\cite{krishna2017visual, hudson2019gqa} using diverse expression templates and functional programs, and automatically obtain the ground-truth annotations at all intermediate steps during the modularized generation process. Furthermore, we carefully balance the dataset by adopting uniform sampling and controlling the distribution of expression-referent pairs over the number of reasoning steps.

In summary, this paper has the following contributions:
\vspace{-0.3cm}

{\flushleft $\bullet$}  A scene graph guided modular neural network is proposed to perform reasoning over a semantic graph and a scene graph using neural modules under the guidance of the linguistic structure of referring expressions, which meets the fundamental requirement of grounding referring expressions.

{\flushleft $\bullet$} A large-scale real-word dataset, Ref-Reasoning, is constructed for grounding referring expressions. Ref-Reasoning includes semantically rich expressions describing objects, attributes, direct and indirect relations with a variety of reasoning layouts.

{\flushleft $\bullet$} Experimental results demonstrate that the proposed method not only significantly surpasses 
existing state-of-the-art algorithms on the new Ref-Reasoning dataset, but also outperforms state-of-the-art structured methods on common benchmark datasets. In addition, it can provide interpretable visual evidences of reasoning.


\begin{figure*}[h]
	\begin{center}
		\includegraphics[width=0.92\linewidth]{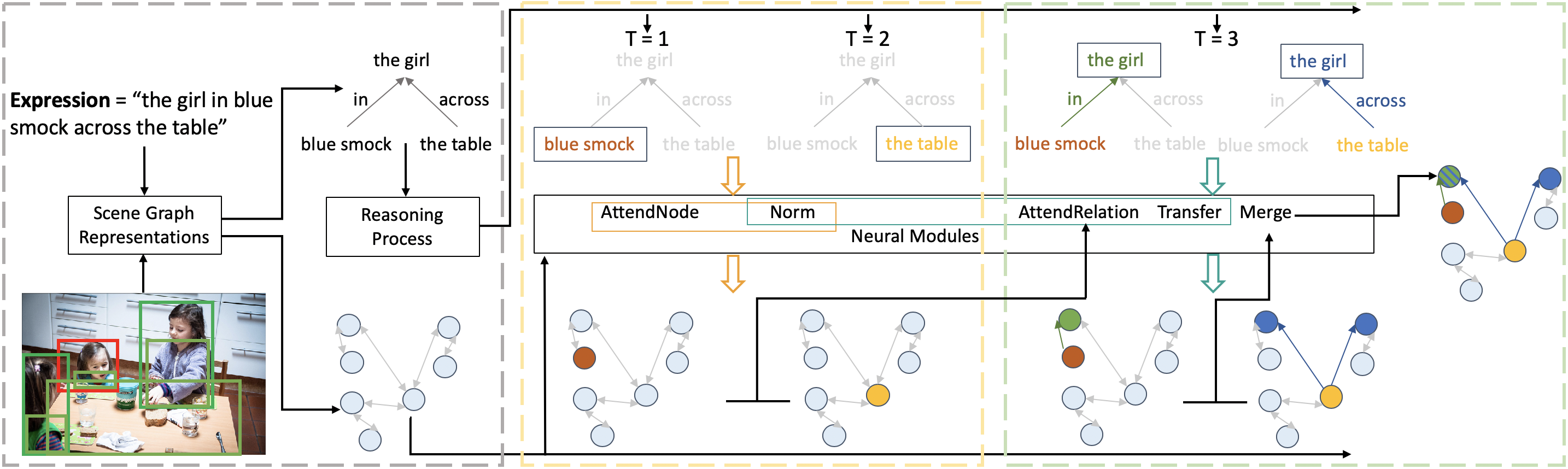}
	\end{center}
	\vspace{-0.2cm}
	\caption{An overview of our Scene Graph guided Modular Network (SGMN)(better viewed in color). Different colors represent different nodes in the language scene graph and their corresponding nodes in the image semantic graph. SGMN parses the expression into a language scene graph and constructs an image semantic graph over the objects in the input image. Next, it performs reasoning under the guidance of the language scene graph. It first locates the nodes in the image semantic graph for the leaf nodes in the language scene graph using neural modules AttendNode and Norm. Then for the intermediate nodes in the language scene graph, it uses AttendRelation, Transfer and Norm modules to attend the nodes in the image semantic graph, and the Merge module to combine the attention results.}
	\label{fig:framework}
\end{figure*}

\section{Related Work}
\subsection{Grounding Referring Expressions}
A referring expression normally not only directly describes the appearance of the referent, but also its relations to other objects in the image, and its reference information depends on the meanings of its constituent expressions and the rules used to compose them \cite{hu2017modeling, yang2019dynamic}. 
However, most of the existing works \cite{zhuang2018parallel, yang2019cross, yang2020relationship} neglect linguistic structures and learn holistic representations for the objects in the image and the expression. 
Recently, there are some works which involve the expression analysis into their models, and learn the components of expression and visual inference from end to end.
The methods in \cite{hu2017modeling, yu2018mattnet, zhang2018grounding} softly decompose the expression into different semantic components relevant to different visual evidences, and compute a matching score for every component. They use fixed semantic components, \eg subject-relation-object triplets~\cite{hu2017modeling} and subject-location-relation components~\cite{yu2018mattnet}, which are not feasible for complex expressions. DGA~\cite{yang2019dynamic} analyzes linguistic structures for complex expressions by iteratively attending their constituent expressions. However, they all resort to self-attention on the expression to explore its linguistic structure but neglect its syntactic information. Another work~\cite{cirik2018using} grounds the referent using a parse tree, where each node of the tree is a word (or phrase) which can be a noun, preposition or verb. 

\subsection{Dataset Bias and Solutions}
Recently, the dataset bias began to be discussed for grounding referring expressions~\cite{cirik2018visual, liu2019clevr}. 
The work in \cite{cirik2018visual} reveals even the linguistically-motivated models tend to learn shallow correlations instead of making use of linguistic structures because of the dataset bias. In addition, expression-independent models can achieve high performance. 
The dataset bias can have a significantly negative impact on the evaluation of a model's readiness for joint understanding and reasoning for language and vision.

In order to address the above problem, the work in \cite{liu2019clevr} proposes a new diagnostic dataset, called CLEVR-Ref+. Same as CLEVR~\cite{johnson2017clevr} in visual question answering, it contains rendered images and automatically generated expressions. In particular, the objects in the images are simple 3D shapes with attributes (\ie, color, size and material), and the expressions are generated using designed templates which include spatial and same-attribute relations. However, the models trained on this synthetic dataset cannot be easily generalized to real-world scenes because the visual contents (\ie, simple 3D shapes with attributes and spatial relations) are too simple to jointly reason about language and vision.

Thanks for the scene graph annotations of real-world images provided in the Visual Genome datasets~\cite{krishna2017visual} and further cleaned in the GQA dataset \cite{hudson2019gqa}, we generate semantically rich expressions over the scene graphs with objects, attributes and relations using carefully designed templates along with functional programs. 

\section{Approach}
We now present the proposed scene graph guided modular network (SGMN). As illustrated in Figure~\ref{fig:framework}, given an input expression and an input image with visual objects, our SGMN first builds a pair of semantic graph and scene graph representations for the image and expression respectively, and then performs structured reasoning over the graphs using neural modules. 

\subsection{Scene Graph Representations}
Scene graph based representations form the basis of our structured reasoning. In particular, the image semantic graph flexibly captures and represents all the visual contents needed for grounding referring expressions in the input image while the language scene graph explores the linguistic structure of the input expression, which defines the layout of the reasoning process. In addition, these two types of graphs have consistent structures, where the nodes and edges of the language scene graph respectively correspond to a subset of the nodes and edges of the image semantic graph.

\vspace{-0.3cm}
\subsubsection{Image Semantic Graph} \label{sec:3_1_1}
Given an image with objects $\mathcal{O} = \{o_i\}_{i=1}^N$ , we define the image semantic graph over the objects $\mathcal{O}$ as a directed graph, $\mathcal{G}^o = (\mathcal{V}^o, \mathcal{E}^o)$, where $\mathcal{V}^o = \{v_i^o\}_{i=1}^N$ is the set of nodes and node $v_i^o$ corresponds to object $o_i$; $\mathcal{E}^o = \{e_{ij}^o\}_{i,j=1}^N$ is the set of directed edges, and $e_{ij}^o$ is the edge from $v_j^o$ to $v_i^o$, which denotes the relation between objects $o_j$ and $o_i$.

For each node $v_i^o$, we obtain two types of features, visual feature $\mathbf{v}_i^o$ extracted from a pretrained CNN model and spatial feature $\mathbf{p}_i^o = [x_i, y_i, w_i, h_i, w_ih_i]$, where $(x_i, y_i)$, $w_i$ and $h_i$ are the normalized top-left coordinates, width and height of the bounding box of node $v_i$ respectively. For each edge $e_{ij}^o$, we compute the edge feature $\mathbf{e}_{ij}^o$ by encoding the relative spatial feature $\mathbf{l}_{ij}^o$ between $v_i^o$ and $v_j^o$ and the visual feature $\mathbf{v}_j^o$ of node $v_j^o$ together because relative spatial information between objects along with their appearance information is the key indicator of their semantic relation~\cite{Dai_2017_CVPR}. Specifically, the relative spatial feature is represented as $\mathbf{l}_{ij}^o = [\frac{x_j-{x_c}_i}{w_i}, \frac{y_j - {y_c}_i}{h_i}, \frac{x_j+w_j-{x_c}_i}{w_i}, \frac{y_j+h_j-{y_c}_i}{h_i}, \frac{w_jh_j}{w_ih_i}]$, where $({x_c}_i, {y_c}_i)$ are the normalized center coordinates of the bounding box of node $v_i^o$. And $\mathbf{e}_{ij}^o$ is the concatenation of an encoded version of $\mathbf{l}_{ij}^o$ and $\mathbf{v}_j^o$, \ie, $\mathbf{e}_{ij}^o = [\mathbf{W}_o^T\mathbf{l}_{ij}^o, \mathbf{v}_j^o]$, where $\mathbf{W}_o$ is a learnable matrix.

\vspace{-0.3cm}
\subsubsection{Language Scene Graph}
Given an expression $S$, we first use an off-the-shelf scene graph parser~\cite{schuster2015generating} to parse the expression into an initial language scene graph, where a node and an edge of the graph correspond to an object and the relation between two objects mentioned in $S$ respectively, and the object is represented as an entity with a set of attributes. 

We define the language scene graph over $S$ as a directed graph $\mathcal{G} = (\mathcal{V}, \mathcal{E})$, where $\mathcal{V} = \{v_m\}_{m=1}^M$ is a set of nodes and node $v_m$ is associated with a noun or noun phrase, which is a sequence of words from $S$; $\mathcal{E} = \{e_k\}_{k=1}^K$ is a set of edges and edge $e_k = ({v_k}_s, r_k, {v_k}_o)$ is a triplet of subject node ${v_k}_s \in \mathcal{V}$, object node ${v_k}_o \in \mathcal{V}$ and relation $r_k$, the direction of which is from ${v_k}_o$ to ${v_k}_s$. Relation $r_k$ is associated with a preposition/verb word or phrase from $S$, and $e_k$ indicates that subject node ${v_k}_s$ is modified by object node ${v_k}_o$.

\subsection{Structured Reasoning}
We perform structured reasoning on the nodes and edges of graphs using neural modules under the guidance of the structure of language scene graph $\mathcal{G}$. In particular, we first design the inference order and reasoning rules for its nodes $\mathcal{V}$ and edges $\mathcal{E}$. Then, we follow the inference order to perform reasoning. For each node, we adopt the AttendNode module to find its corresponding node in graph $\mathcal{G}^o$ or use the Merge module to combine information from its incident edges. For each edge, we execute specific reasoning steps using carefully designed neural modules, including AttendNode, AttendRelation and Transfer.

\vspace{-0.3cm}
\subsubsection{Reasoning Process}
In this section, we first introduce the inference order, and then present specific reasoning steps on the nodes and edges respectively. In general, for every node in language scene graph $\mathcal{G}$, we learn its attention map over the nodes of image semantic graph $\mathcal{G}^o$ on the basis of its connections.

Given a language scene graph $\mathcal{G}$, we locate the node with zero out-degree as its referent node $v_{ref}$ because the referent is usually modified by other entities rather than modifying other entities in a referring expression. Then, we perform breadth-first traversal of the nodes in graph $\mathcal{G}$ from the referent node $v_{ref}$ by reversing the direction of all edges, meanwhile, push the visited nodes into a stack which is initially empty. Next, we iteratively pop one node from the stack and perform reasoning on the popped node. The stack determines the inference order for the nodes, and one node can reach the top of the stack only after all of its modifying nodes have been processed. This inference order essentially converts graph $\mathcal{G}$ into a directed acyclic graph. Without loss of generality, suppose node $v_m$ is popped from the stack in the present iteration, and we carry out reasoning on node $v_m$ on the basis of its connections to other nodes. There are two different situations: 1) If the in-degree of $v_m$ is zero, $v_m$ is a leaf node, which means node $v_m$ is not modified by any other nodes. Thus, node $v_m$ should be associated with the nodes of image semantic graph $\mathcal{G}^o$ independently; 2) otherwise, if node $v_m$ has incident edges $\mathcal{E}_m \in \mathcal{E}$ starting from other nodes, $v_m$ is an intermediate node, and its attention map over $\mathcal{V}^o$ should depend on the attention maps of its connected nodes and the edges between them.

{\flushleft \bf Leaf node}. We learn an embedding for the words associated with the nodes of the language scene graph $\mathcal{G}$ in advance. Then, for node $v_m$, suppose its associated phrase consists of words $\{w_t\}_{t=1}^T$, and the embedded feature vectors for these words are $\{\mathbf{f}_t\}_{t=1}^T$. We use a bi-directional LSTM~\cite{hochreiter1997long} to compute the context of every word in this phrase, and define the concatenation of the forward and backward hidden vectors of a word $w_t$ as its context, denoted as $\mathbf{h}_t$. Meanwhile, we represent the whole phrase using the concatenation of the last hidden vectors of both directions, denoted as $\mathbf{h}$. In a referring expression, an individual entity is often described by its appearance and spatial location. Therefore, we learn feature representations for node $v_m$ from both appearance and spatial location. In particular, inspired by self-attention in \cite{hu2017modeling, yang2019cross, yu2018mattnet}, we first learn the attention over each word on the basis of its context, and obtain feature representations $\mathbf{v}^{look}_m$ and $\mathbf{v}^{loc}_m$ at node $v_m$ by aggregating attention weighted word embedding as follows,
\begin{equation}
\begin{aligned}
	\alpha^{look}_{t, m} &= \frac{\text{exp}(\mathbf{W}_{look}^T\mathbf{h}_t)}{\sum_{t=1}^T\text{exp}(\mathbf{W}_{look}^T\mathbf{h}_t)}, \mathbf{v}^{look} _m= \sum_{t=1}^T\alpha^{look}_{t,m}\mathbf{f}_t\\
	\alpha^{loc}_{t, m} &= \frac{\text{exp}(\mathbf{W}_{loc}^T\mathbf{h}_t)}{\sum_{t=1}^T\text{exp}(\mathbf{W}_{loc}^T\mathbf{h}_t)},  \mathbf{v}^{loc} _m= \sum_{t=1}^T\alpha^{loc}_{t,m}\mathbf{f}_t,
\end{aligned}
\end{equation}
where $\mathbf{W}_{look}$ and $\mathbf{W}_{loc}$ are learnable parameters, and $\mathbf{v}^{look}_m$ and $\mathbf{v}^{loc}_m$ correspond to the appearance and spatial location of node $v_m$. Then, we feed these two features into the \textbf{AttendNode} neural module to compute attention maps $\{\lambda^{look}_{n, m}\}_{n=1}^N$ and $\{\lambda^{loc}_{n, m}\}_{n=1}^N$ over the nodes of image semantic graph $\mathcal{G}^{o}$. Finally, we combine these two attention maps to obtain the final attention map for node $v_m$. A noun phrase may place emphasis on appearance, spatial location or both of them. We flexibly adapt to the variations of noun phrases by learning a pair of weights at node $v_m$ for the attention maps related to appearance and spatial location. The weights (\ie $\beta^{look}$ and $\beta^{loc}$) and the final attention map $\{\lambda_{n,m}\}_{n=1}^N$ for node $v_m$ are computed as follows,
\begin{equation}
\begin{aligned}
\beta^{look} &= \text{sigmoid}(\mathbf{W}^T_0\mathbf{h} + b_0) \\
\beta^{loc} &= \text{sigmoid}(\mathbf{W}^T_1\mathbf{h} + b_1) \\
\lambda_{n,m} &= \beta^{look}\lambda_{n,m}^{look} + \beta^{loc}\lambda_{n,m}^{loc} \\
\{\lambda_{n,m}\}_{n=1}^N &= \text{Norm}(\{\lambda_{n,m}\}_{n=1}^N),
\end{aligned}
\end{equation}
where $\mathbf{W}^T_0$, $b_0$, $\mathbf{W}^T_1$ and $b_1$ are learnable parameters, and the \textbf{Norm} module is used to constrain the scale of the attention map.
\vspace{-2mm}

{\flushleft \bf Intermediate node}. As an intermediate node, $v_m$ is connected to other nodes that modify it, and such connections are actually a subset of edges, $\mathcal{E}_m \in \mathcal{E}$, incident to $v_m$. We compute an attention map over the edges of image semantic graph $\mathcal{G}^o$ for each edge in this subset, then transfer and combine all these attention maps to obtain a final attention map for node $v_m$.

For each edge $e_k = ({v_k}_s, r_k, {v_k}_o)$ in $\mathcal{E}_m$ (where ${v_k}_s$ is exactly $v_m$), we first form a sentence associated with $e_k$ by concatenating the words or phrases associated with ${v_k}_s$, $r_k$ and ${v_k}_o$. Then, we obtain the embedded feature vectors $\{\mathbf{f}_t\}_{t=1}^T$ and word contexts $\{\mathbf{h}_t\}_{t=1}^T$ for the words $\{w_t\}_{t=1}^T$ in this sentence and the feature representation of the whole sentence by following the same computation for leaf nodes. Next, we compute the attention map for node ${v_k}_s$ from two different aspects, \ie subject description and relation-based transfer, because $e_k$ not only directly describes subject ${v_k}_s$ itself but also its relation to object ${v_k}_o$. From the aspect of subject description, same as the computation for leaf nodes, we obtain attention maps corresponding to the appearance and spatial location of ${v_k}_s$ ( \ie $\{\lambda^{look}_{n, k_s}\}_{n=1}^N$ and $\{\lambda^{loc}_{n, k_s}\}_{n=1}^N$) and weights (\ie $\beta^{look}_{k_s}$ and $\beta^{loc}_{k_s}$) to combine them.
From the aspect of relation-based transfer, we first compute a relational feature representation for edge $e_k$ as follows,
\begin{equation}
	\alpha^{rel}_{t, k} = \frac{\text{exp}(\mathbf{W}_{rel}^T\mathbf{h}_t)}{\sum_{t=1}^T\text{exp}(\mathbf{W}_{rel}^T\mathbf{h}_t)}, \mathbf{r} _k= \sum_{t=1}^T\alpha^{rel}_{t,k}\mathbf{f}_t\\
\end{equation}
where $\mathbf{W}_{rel}$ is a learnable parameter. Then we feed the relational representation $\mathbf{r}_k$ to the \textbf{AttendRelation} neural module to attend the relation $\mathbf{r}_k$ over the edges $\mathcal{E}_{ij}^o$ of graph $\mathcal{G}^o$, and the computed attention weights are denoted as $\{\gamma_{ij, k}\}_{i,j=1}^N$. Moreover, we use the \textbf{Transfer} module and the \textbf{Norm} module to transfer the attention map $\{\lambda_{n, k_o}\}_{n=1}^N$ for object node ${v_k}_o$ to node $v_m$ by modulating $\{\lambda_{n, k_o}\}_{n=1}^N$ with the attention weights on edges $\{\gamma_{ij, k}\}_{i,j=1}^N$, and the transferred attention map for node $v_m$ is denoted as $\{\lambda^{rel}_{n, k_s}\}_{n=1}^N$. It is worth mentioning that object node ${v_k}_o$ has been accessed before and the attention map $\{\lambda_{n, k_o}\}_{n=1}^N$ for node ${v_k}_o$ has been computed. Next, we estimate the weight of relation at edge $e_k$ and integrate the attention maps for node ${v_k}_s$ related to subject description and relation-based transfer to obtain attention map  $\{\lambda_{n, k_s}\}_{n=1}^N$ for node ${v_k}_s$ contributed by edge $e_k$, and $\{\lambda_{n, k_s}\}_{n=1}^N$ is defined as follows.
\begin{equation}
	\begin{aligned}
	\beta^{rel}_k &= \text{sigmoid}(\mathbf{W}^T_2\mathbf{h} + b_2) \\
	\lambda_{n, k_s} &= \beta^{look}_{k_s}\lambda^{look}_{n, k_s} + \beta^{loc}_{k_s}\lambda^{loc}_{n, k_s}+\beta^{rel}_{n, k_s}\lambda^{rel}_{n, k_s} \\
	\{\lambda_{n, k_s}\}_{n=1}^N &= \text{Norm}(\{\lambda_{n, k_s}\}_{n=1}^N),
	\end{aligned}
\end{equation}
where $\mathbf{W}_2$ and $b_2$ are learnable parameters.

Finally, we combine the attention maps $\{\{\lambda_{n, k_s}\}_{n=1}^N\}$ for node $v_m$ contributed by all edges in $\mathcal{E}_m$ using the \textbf{Merge} module followed by the \textbf{Norm} module to obtain the final attention map $\{\lambda_{n,m}\}_{n=1}^N$ for node $v_m$.

\vspace{-0.3cm}
\subsubsection{Neural Modules}
We present a series of neural modules to perform specific reasoning steps, inspired by the neural modules in \cite{shi2019explainable}. In particular, the AttendNode and AttendRelation modules are used to connect the language mode with the vision mode. They receive feature representations of linguistic contents from the language scene graph and output attention maps of the features defined over visual contents in the image semantic graph. The Merge, Norm and Transfer modules are adopted to further integrate and transfer attention maps over the nodes and edges of the image semantic graph.
\vspace{-2mm}

{\flushleft \bf AttendNode [appearance query, location query]} module aims to find relevant nodes among the nodes of the image semantic graph $\mathcal{G}^o$ given an appearance query and location query. It takes the query vectors of the appearance query and location query as inputs and generate attention maps $\{\lambda^{look}_n\}_{n=1}^N$ and $\{\lambda^{loc}_n\}_{n=1}^N$ over the nodes $\mathcal{V}^o$, where every node $v^o_n \in \mathcal{V}^o$ has two attention weights, \ie, $\lambda^{look}_n \in [-1, 1]$ and $ \lambda^{loc}_n \in [-1, 1]$. The query vectors are linguistic features at nodes of the language scene graph, denoted as $\mathbf{v}^{look}$ and $\mathbf{v}^{loc}$. For node $v^o_n$ in graph $\mathcal{G}^o$, its attention weights $\lambda^{look}_n$ and $ \lambda^{loc}_n$ are defined as follows,
\begin{equation}
\begin{aligned}
\lambda^{look}_n &= \langle\text{L2Norm}(\text{MLP}_0(\mathbf{v}^o_n)), \text{L2Norm}(\text{MLP}_1(\mathbf{v}^{look}))\rangle, \\
\lambda^{loc}_n &=  \langle\text{L2Norm}(\text{MLP}_2(\mathbf{p}^o_n)), \text{L2Norm}(\text{MLP}_3(\mathbf{v}^{loc}))\rangle,
\end{aligned}
\end{equation}
where $\text{MLP}_0()$, $\text{MLP}_1()$, $\text{MLP}_2()$ and $\text{MLP}_3()$ are multilayer perceptrons consisting of several linear and $\text{ReLU}$ layers, $\text{L2Norm()}$ is the L2 normalization, and $\mathbf{v}^o_n$ and $\mathbf{p}_n^o$ are the visual feature and spatial feature at node $v^o_n$ respectively, which are mentioned in Section~\ref{sec:3_1_1}.
\vspace{-2mm}

{\flushleft \bf AttendRelation [relation query]} module aims to find relevant edges in the image semantic graph $\mathcal{G}^o$ given a relation query. The purpose of a relation query is to establish connections between nodes in graph $\mathcal{G}^o$. Given query vector $\mathbf{e}$, the attention weights $\{\gamma_{ij}\}_{i,j=1}^N$ on edges $\{\mathbf{e}^o_{ij}\}_{i,j=1}^N$ are defined as follows,
\begin{equation}
	\gamma_{ij} = \sigma(  \langle\text{L2Norm}(\text{MLP}_5(\mathbf{e}^o_{ij})), \text{L2Norm}(\text{MLP}_1(\mathbf{e}))\rangle)
\end{equation}
where $\text{MLP}_5()$, $\text{MLP}_5()$ are multilayer perceptrons, and the $\text{ReLU}$ activation function $\sigma$ ensures the attention weights are larger than zero.
\vspace{-2mm}

{\flushleft \bf Transfer} module aims to find new nodes by passing attention weights $\{\lambda_n\}_{n=1}^N$ on nodes that modify those new nodes along attended edges $\{\gamma_{ij}\}_{i,j=1}^N$. The updated attention weights $\{\lambda^{new}_n\}_{n=1}^N$ are calculated as follows, \vspace{-2mm}
\begin{equation}
\begin{aligned}
\lambda^{new}_n &= \sum_{j=1}^N\gamma_{n,j}\lambda_{j}.
\end{aligned}
\end{equation}

{\flushleft \bf Merge} module aims to combine multiple attention maps generated from different edges of the same node, where the attention weights over edges are computed individually. Given the set of attention maps $\Lambda$ for a node, the merged attention map $\{\lambda_{n}\}_{n=1}^N$ is defined as follows,
\begin{equation}
	\lambda_{n} = \sum_{\{\lambda'_n\}_{n=1}^N \in \Lambda}\lambda'_{n}.
\end{equation}
\vspace{-4mm}

{\flushleft \bf Norm} module aims to set the range of weights in attention maps to $[-1,\, 1]$. If the maximum absolute value of an attention map is larger than $1$, the attention map is divided by the maximum absolute value.

\subsection{Loss Function}
Once all the nodes in the stack have been processed, the final attention map for the referent node of the language scene graph is obtained. This attention map is denoted as $\{\lambda_{n, ref}\}_{n=1}^N$.
As in previous methods for grounding referring expressions~\cite{hu2017modeling}, during the training phase, we adopt the cross-entropy loss, which is defined as
\begin{equation}
\begin{aligned}
p_{i} = \text{exp}(\lambda_{i, ref}) / {\sum_{n=1}^N\text{exp}(\lambda_{n, ref})}, 
\text{loss} = -\text{log}(p_{gt})
\end{aligned}
\end{equation}
where $p_{gt}$ is the probability of the ground-truth object. During the inference phase, we predict the referent by choosing the object with the highest probability.

\section{Ref-Reasoning Dataset}
The proposed dataset is built on the scenes from the GQA dataset~\cite{hudson2019gqa}. 
We automatically generate referring expressions for every image on the basis of the image scene graph using a diverse set of expression templates. 

\subsection{Preparation}
{\flushleft \bf Scene Graph}. We generate referring expressions according to the ground-truth image scene graphs. 
Specifically, we adopt the scene graph annotations provided by the Visual Genome dataset~\cite{krishna2017visual} and further normalized by the GQA dataset.  In a scene graph annotation of an image, each node represents an object with about 1-3 attributes, and each edge represents a relation (\ie, semantic relation, spatial relation and comparatives) between two objects. In order to use the scene graphs for referring expression generation, we remove some unnatural edges and classes, 
\eg, ``nose left of eyes". In addition, we add edges between objects to represent same-attribute relations between objects, \ie, ``same material", ``same color" and ``same shape''. In total, there are 1,664 object classes, 308 relation classes and 610 attribute classes in the adopted scene graphs.
\vspace{-2mm}

{\flushleft \bf Expression Template}. In order to generate referring expressions with diverse reasoning layouts, 
for each specified number of nodes, we design a family of referring expression templates for each reasoning layout. 
We generate expressions according to layouts and templates using functional programs, and the functional program for each template can be easily obtained according to the layout. In particular, layouts are sub-graphs of directed acyclic graphs, where only one node (\ie, the root node) has zero out-degree and other nodes can reach the root node. The functional program for a layout provides a step-wise plan for reaching the root node from leaf nodes (\ie, the nodes with zero in-degree) by traversing all the nodes and edges in this layout, and templates are parameterized natural language expressions, where the parameters can be filled in. 
Moreover, we set the constraint that the number of nodes in a template ranges from one to five.


\begin{table*}[h!]
	\begin{center}
		\resizebox{0.53\textwidth}{!}
		{
			\begin{tabular}{|l|c|c|c|c|c|c|}\hline
				&\multicolumn{4}{|c|}{Number of Objects} & \multicolumn{2}{|c|}{Split} \\ \hline
				& one & two & three & $>=$ four  & val & test \\ \hline
				CNN & 10.57 & 13.11 & 14.21 & 11.32 & 12.36 & 12.15 \\
				CNN+LSTM & 75.29 & 51.85 & 46.26 & 32.45 & 42.38 & 42.43 \\
				DGA  & 73.14 & 54.63 & 48.48 & 37.63 & 45.37 & 45.87 \\
				CMRIN & 79.20 & 56.87 & 50.07 & 35.29 & 45.43 & 45.87  \\
				Ours SGMN & \textbf{79.71} & \textbf{61.77} & \textbf{55.57} & \textbf{41.89} & \textbf{51.04} & \textbf{51.39} \\ \hline
				CMRIN*  & 79.83 & 58.02 & 51.51 & 37.65 & 47.40 & 47.69 \\
				DGA*$^{\ddagger}$ & 78.57 & 59.85 & 53.37 & 40.03 & 48.95 & 49.51  \\
				Ours SGMN* & \textbf{80.17} & \textbf{62.24} & \textbf{56.24} & \textbf{42.45} & \textbf{51.59} & \textbf{51.95} \\ \hline
			\end{tabular}
		}
	\end{center}
	\caption{Comparison with baselines and state-of-the-art methods on Ref-Reasoning dataset. We use * to indicate that this model uses bottom-up features and use $^{\ddagger}$ to indicate that the implementation of this model is slightly different from it in origin paper. The best performing method is marked in bold.}
	\label{tab:reasoning}
	\vspace{-0.3cm}
\end{table*}

\subsection{Generation Process}
Given an image, we generate dozens of expressions from the scene graph of the image, and the generation process for one expression is summarized as follows,
\begin{itemize}
	\setlength{\itemsep}{0pt}
	\setlength{\parsep}{0pt}
	\setlength{\parskip}{0pt}
	\item Randomly sample the referent node and randomly decide the number of nodes, denoted as $C$.
	\item Randomly sample a sub-graph with $C$ nodes including the referent node in the scene graph.
	\item Judge the layout of the sub-graph and randomly sample a referring expression template from the family of templates corresponding to the layout.
	\item Fill in the parameters in the template using contents of the sub-graph, including relations and objects with randomly sampled attributes.
	\item Execute the functional program with filled parameters and accept the expression if the referred object is unique in the scene graph.
\end{itemize}

Note that we perform extra operations during the generation process: 1) If there are objects that have same-attribute relations in the sub-graph, we avoid choosing the attributes that appear in such relations for these objects. This restriction intends to make the modified node identified by the relation edge instead of the attribute directly. 2) To help balance the dataset, during the process of random sampling, we decrease the chances of nodes and relations whose classes most commonly exist in the scene graphs. In addition, we increase the chances of multi-order relationships with $C=3$ or $C=4$ to reasonably increase the level of difficulty for reasoning. 3) We define a difficulty level for a referring expression. We find its shortest sub-expression which can identify the referent in the scene graph, and the number of objects in the sub-expression is defined as the difficulty level. For example, if there is only one bottle in an image, the difficulty level of ``the bottle on a table beside a plate'' is still one even though it describes three objects and their relations. Then, we obtain the balanced dataset and its final splits by randomly sampling expressions of images according to their difficulty level and the number of nodes described by them.

\section{Experiments}
\subsection{Datasets}
We have conducted extensive experiments on the proposed Ref-Reasoning dataset as well as on three commonly used benchmark datasets (\ie, RefCOCO\cite{yu2016modeling}, RefCOCO+\cite{yu2016modeling} and RefCOCOg\cite{mao2016generation}).
Ref-Reasoning 
contains 791,956 referring expressions in 83,989 images. It has 721,164, 36,183 and 34,609 expression-referent pairs for training, validation and testing, respectively. 
Ref-Reasoning includes semantically rich expressions describing objects, attributes, direct relations and indirect relations with different layouts. 
RefCOCO and RefCOCO+ datasets includes short expressions collected from an interactive game interface. RefCOCOg collects from a non-interactive settings and it has longer complex expressions.

\subsection{Implementation and Evaluation}
The performance of grounding referring expressions is evaluated by accuracy, \ie, the fraction of correct predictions of referents.

For the Ref-Reasoning dataset, we use a ResNet-101 based Faster R-CNN \cite{ren2015faster, he2016deep} as the backbone, and adopt a feature extractor which is trained on the training set of GQA with an extra attribute loss following~\cite{Anderson2017up-down}. Visual features of annotated objects are extracted from the pool5 layer of the feature extractor. For the three common benchmark datasets (\ie, RefCOCO, RefCOCO+ and RefCOCOg), we follow CMRIN~\cite{yang2019cross} to extract the visual features of objects in images. 
To keep the image semantic graph sparse and reduce computational cost, we connect each node in the image semantic graph to its five nearest nodes based on the distances between their normalized center coordinates. 
We set the mini-batch size to 64. All the models are trained by the Adam optimizer~\cite{kingma2014adam} with the learning rate set to 0.0001 and 0.0005 for the Ref-Reasoning dataset and other benchmark datasets respectively.

\subsection{Comparison with the State of the Art}
We conduct experimental comparisons between the proposed SGMN and existing state-of-the-art methods on both the collected Ref-Reasoning dataset and three commonly used benchmark datasets. 
\vspace{-2mm}

{\flushleft \bf Ref-Reasoning Dataset.} We evaluate two baselines (\ie, a CNN model and a CNN+LSTM model), two state-of-the-art methods (\ie, CMRIN~\cite{yang2019cross} and DGA~\cite{yang2019dynamic}) and the proposed SGMN on the Ref-Reasoning dataset. The CNN model is allowed to access objects and images only. 
The CNN+LSTM model embeds objects and expressions into a common feature space and learns matching scores between them. 
For CMRIN and DGA, we adopt their default settings~\cite{yang2019cross, yang2019dynamic} in our evaluation. For a fair comparison, all the models use the same visual object features and the same setting in LSTMs.

Table~\ref{tab:reasoning} shows the evaluation results on the Ref-Reasoning dataset. The proposed SGMN significantly outperforms the baselines and existing state-of-the-art models, and it consistently achieves the best performance on all the splits of the testing set, where different splits need different numbers of reasoning steps. 
The CNN model has a low accuracy of 12.15\%, which is much lower than the accuracy (\ie, 41.1\% \cite{cirik2018visual}) of the image-only model for the RefCOCOg dataset, which demonstrates that joint understanding of images and text is required on Ref-Reasoning. The CNN+LSTM model achieves a high accuracy of 75.29\% on the split where expressions directly describe the referents. This is because relation reasoning is not required in this split and LSTM may be qualified to capture the semantics of expressions. Compared with the CNN+LSTM model, DGA and CMRIN achieve higher performance on the two-, three- and four-node splits because they learn a language-guided contextual representation for objects.
\vspace{-2mm}
\begin{table}[h]
	\begin{center}
		\resizebox{0.47\textwidth}{!}
		{
			\begin{tabular}{|l|c|c|c|c|c|}\hline
				&\multicolumn{2}{|c|}{RefCOCO} & \multicolumn{2}{|c|}{RefCOCO+} & \multicolumn{1}{|c|}{RefCOCOg} \\ \hline
				& testA & testB  & testA & testB & test \\ \hline
				\textbf{Holistic Models} &&&&& \\
				CMN \cite{hu2017modeling} & 75.94 & 79.57 & 59.29 & 59.34 & - \\
				ParallelAttn \cite{zhuang2018parallel} & 80.81 & 81.32  & 66.31 & 61.46 & -  \\
				MAttNet* \cite{yu2018mattnet} & 85.26 & 84.57 & 75.13 & 66.17 & 78.12  \\
				CMRIN* \cite{yang2019cross} & \textbf{87.63} & 84.73 & \textbf{80.93} & 68.99 & 80.66 \\
				DGA* \cite{yang2019dynamic} & 86.64 & 84.79 & 78.31 & 68.15 & 80.26  \\ \hline
				\textbf{Structured Models} & &&&& \\
				MattNet* + parser \cite{yu2018mattnet}  & 79.71 & 81.22 & 68.30 & 62.94 & 73.72 \\
				RvG-Tree* \cite{hong2019learning} & 82.52 & 82.90 & 70.21 & 65.49 & 75.20 \\
				DGA* + parser \cite{yang2019dynamic} & 84.69 & 83.69 & 74.83 & 65.43 & 76.33 \\
				NMTree* \cite{liu2019learning} & 85.63 & 85.08 & 75.74 & 67.62 & 78.21 \\
				MSGL* \cite{liu2019referring} & 85.45 & 85.12 & 75.31 & 67.50 & 78.46 \\
				Ours SGMN*  & 86.67 & \textbf{85.36} & 78.66 & \textbf{69.77} & \textbf{81.42} \\ \hline
			\end{tabular}
		}
		\vspace{0.15cm}
		\caption{Comparison with state-of-the-art methods on RefCOCO, RefCOCO+ and RefCOCOg. We use * to indicate that this model uses resnet101 features. None-superscript indicates that model uses vgg16 features. The best performing method is marked in bold. }
		\label{tab:common}
		\vspace{-0.7cm}
	\end{center}
\end{table}

{\flushleft \bf Common Benchmark Datasets.} Quantitative evaluation results on RefCOCO, RefCOCO+ and RefCOCOg datasets are shown in Table~\ref{tab:common}. The proposed SGMN consistently outperforms existing structured methods across all the datasets, and it improves the average accuracy over the testing sets achieved by the best performing existing structured method by 0.92\%, 2.54\% and 2.96\% respectively on the RefCOCO, RefCOCO+ and RefCOCOg datasets.
Moreover, it also surpasses all the existing models on the RefCOCOg dataset which has relatively longer complex expressions with an average length 8.43, and achieves a performance comparable to the best performing holistic method on the other two common benchmark datasets. Note that holistic models usually have higher performance than structured models on the common benchmark datasets because those datasets include many simple expressions describing the referents without relations, and holistic models are prone to learn shallow correlations without reasoning and may exploit this dataset bias~\cite{ cirik2018visual, liu2019clevr}. In addition, the inference mechanism of holistic methods has poor interpretability.

\subsection{Qualitative Evaluation}
\vspace{-0.3cm}
\begin{figure}[h]
	\begin{center}
		\includegraphics[width=1.0\linewidth]{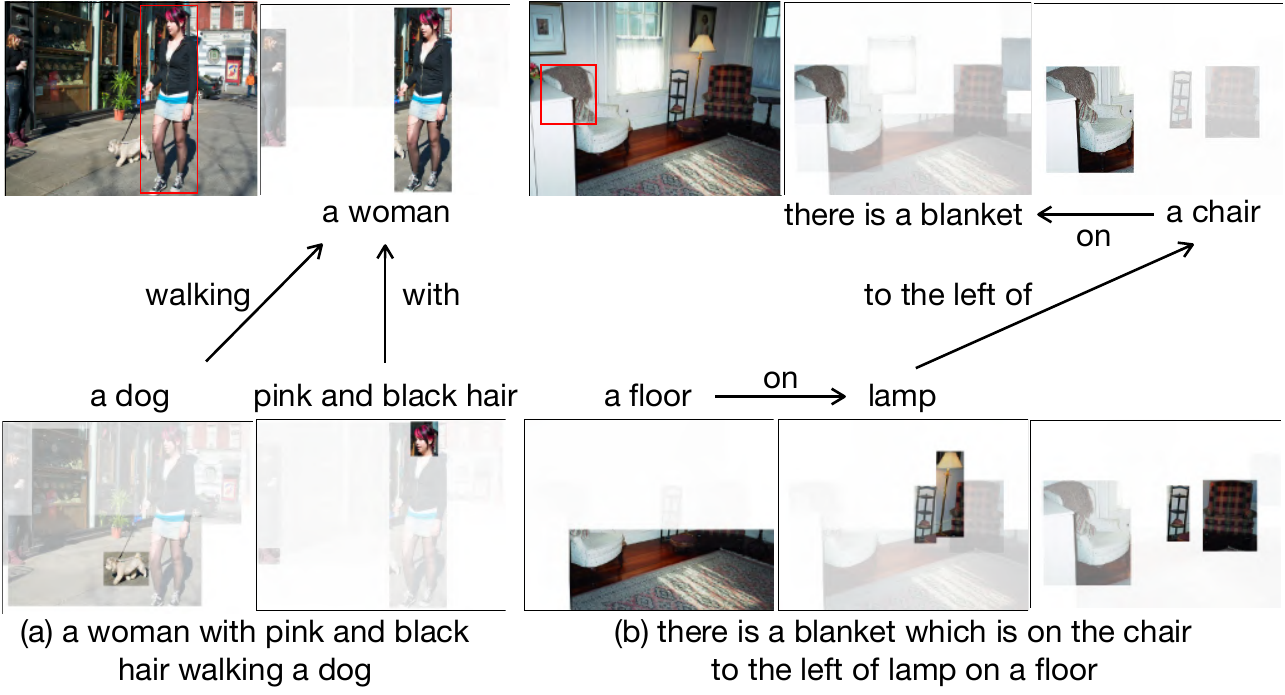}
	\end{center}
	\vspace{-0.2cm}
	\caption{Qualitative results showing the attention maps over the objects along the language scene graphs predicted by the SGMN.}
	\label{fig:vis}
\end{figure}

Visualizations of two examples along with their language scene graphs and attention maps over the objects in images at every node of the language scene graphs are shown in Figure~\ref{fig:vis}. 
This qualitative evaluation results demonstrate that the proposed SGMN can generate interpretable visual evidences of intermediate steps in the reasoning process. 
In Figure~\ref{fig:vis}(a), SGMN parses the expression into a tree structure and finds the referred ``woman'' who is walking ``a dog'' and meanwhile is with ``pink and black hair''. Figure~\ref{fig:vis}(b) shows a more complex expression which describes four objects and their relations. SGMN first successfully changes from the initial attention map (bottom-right) to the final attention map (top-right) at the node ``a chair''  by performing relational reasoning along the edges (\ie, triplets (``a chair'', ``to the left of'' , ``lamp'') and (``lamp'', ``on'', ``a floor'')), and then identifies the target ``blanket'' on that chair.

\subsection{Ablation Study}

\vspace{-0.3cm}
\begin{table}[h!]
	\begin{center}
		\resizebox{0.47\textwidth}{!}
		{
			\begin{tabular}{|l|c|c|c|c|c|c|}\hline
				&\multicolumn{4}{|c|}{Number of Objects} & \multicolumn{2}{|c|}{Split} \\ \hline
				& one & two & three & $>=$ four  & val & test \\ \hline
				w/o transfer & 79.14 & 48.51 & 45.97 & 31.57 & 40.66 & 41.88 \\
				w/o norm & 79.37 & 49.44 & 45.61 & 31.57 & 40.80 & 41.93 \\
				max merge & 78.71 & 54.00 & 50.34  & 34.76 & 44.50 & 45.27 \\
				min merge & 78.83  & 53.83 & 51.11 & 35.79 & 45.25 & 46.00  \\
				Ours SGMN & \textbf{79.71} & \textbf{61.77} & \textbf{55.57} & \textbf{41.89} & \textbf{51.04} & \textbf{51.39} \\ \hline
			\end{tabular}
		}
	\end{center}
	\caption{Ablation study on Ref-Reasoning dataset. The best performing method is marked in bold.}
	\label{tab:ablation}
	\vspace{-0.1cm}
\end{table}

To demonstrate the effectiveness of reasoning under the guidance of scene graphs inferred from referring expressions as well as the design of neural modules, we train four additional models for comparison. The results are shown in Table~\ref{tab:ablation}. All the models have similar performance on the split of expressions directly describing the referents. For the other splits, SGMN without the Transfer module and SGMN without the Norm module have much lower performance than the original SGMN because the former treats the referent as an isolated node without performing relation reasoning while the latter unfairly treats different relational edges and the nodes connected by them. Next, we explore different options of the function (\ie, max, min and sum) used in the Merge module. Compared to SGMN with sum-merge, its performance with min-merge and max-merge drops because max-merge only captures the most significant relation for each intermediate node and min-merge is sensitive to parsing errors and recognition errors.


\section{Conclusion}
In this paper, we present a scene graph guided modular network (SGMN) for grounding referring expressions. It performs graph-structured reasoning over the constructed graph representations of the input image and expression using neural modules. In addition, we propose a large-scale real-world dataset for structured referring expression reasoning, named Ref-Reasoning. Experimental results demonstrate that SGMN not only significantly outperforms existing state-of-the-art algorithms on the new Ref-Reasoning dataset, but also surpasses state-of-the-art structured methods on commonly used benchmark datasets. Moreover, it can generate interpretable visual evidences of reasoning via a graph attention mechanism.

{\small
\bibliographystyle{ieee_fullname}
\bibliography{gre_reason}

\begin{thebibliography}{10}\itemsep=-1pt

\bibitem{Anderson2017up-down}
Peter Anderson, Xiaodong He, Chris Buehler, Damien Teney, Mark Johnson, Stephen
  Gould, and Lei Zhang.
\newblock Bottom-up and top-down attention for image captioning and visual
  question answering.
\newblock In {\em Proceedings of the IEEE conference on computer vision and
  pattern recognition (CVPR)}, 2018.

\bibitem{andreas2016neural}
Jacob Andreas, Marcus Rohrbach, Trevor Darrell, and Dan Klein.
\newblock Neural module networks.
\newblock In {\em Proceedings of the IEEE Conference on Computer Vision and
  Pattern Recognition}, pages 39--48, 2016.

\bibitem{cirik2018using}
Volkan Cirik, Taylor Berg-Kirkpatrick, and Louis-Philippe Morency.
\newblock Using syntax to ground referring expressions in natural images.
\newblock In {\em Thirty-Second AAAI Conference on Artificial Intelligence},
  2018.

\bibitem{cirik2018visual}
Volkan Cirik, Louis-Philippe Morency, and Taylor Berg-Kirkpatrick.
\newblock Visual referring expression recognition: What do systems actually
  learn?
\newblock {\em arXiv preprint arXiv:1805.11818}, 2018.

\bibitem{Dai_2017_CVPR}
Bo Dai, Yuqi Zhang, and Dahua Lin.
\newblock Detecting visual relationships with deep relational networks.
\newblock In {\em The IEEE Conference on Computer Vision and Pattern
  Recognition (CVPR)}, July 2017.

\bibitem{he2016deep}
Kaiming He, Xiangyu Zhang, Shaoqing Ren, and Jian Sun.
\newblock Deep residual learning for image recognition.
\newblock In {\em Proceedings of the IEEE conference on computer vision and
  pattern recognition (CVPR)}, pages 770--778, 2016.

\bibitem{hochreiter1997long}
Sepp Hochreiter and J{\"u}rgen Schmidhuber.
\newblock Long short-term memory.
\newblock {\em Neural computation}, 9(8):1735--1780, 1997.

\bibitem{hong2019learning}
R. {Hong}, D. {Liu}, X. {Mo}, X. {He}, and H. {Zhang}.
\newblock Learning to compose and reason with language tree structures for
  visual grounding.
\newblock {\em IEEE Transactions on Pattern Analysis and Machine Intelligence},
  pages 1--1, 2019.

\bibitem{hu2017modeling}
Ronghang Hu, Marcus Rohrbach, Jacob Andreas, Trevor Darrell, and Kate Saenko.
\newblock Modeling relationships in referential expressions with compositional
  modular networks.
\newblock In {\em Proceedings of the IEEE Conference on Computer Vision and
  Pattern Recognition}, pages 1115--1124, 2017.

\bibitem{hudson2019gqa}
Drew~A Hudson and Christopher~D Manning.
\newblock Gqa: A new dataset for real-world visual reasoning and compositional
  question answering.
\newblock In {\em Proceedings of the IEEE Conference on Computer Vision and
  Pattern Recognition}, pages 6700--6709, 2019.

\bibitem{johnson2017clevr}
Justin Johnson, Bharath Hariharan, Laurens van~der Maaten, Li Fei-Fei, C
  Lawrence~Zitnick, and Ross Girshick.
\newblock Clevr: A diagnostic dataset for compositional language and elementary
  visual reasoning.
\newblock In {\em Proceedings of the IEEE Conference on Computer Vision and
  Pattern Recognition}, pages 2901--2910, 2017.

\bibitem{kazemzadeh2014referitgame}
Sahar Kazemzadeh, Vicente Ordonez, Mark Matten, and Tamara Berg.
\newblock Referitgame: Referring to objects in photographs of natural scenes.
\newblock In {\em Proceedings of the 2014 conference on empirical methods in
  natural language processing (EMNLP)}, pages 787--798, 2014.

\bibitem{kingma2014adam}
Diederik~P Kingma and Jimmy Ba.
\newblock Adam: A method for stochastic optimization.
\newblock {\em International Conference on Learning Representations}, 2015.

\bibitem{krishna2017visual}
Ranjay Krishna, Yuke Zhu, Oliver Groth, Justin Johnson, Kenji Hata, Joshua
  Kravitz, Stephanie Chen, Yannis Kalantidis, Li-Jia Li, David~A Shamma, et~al.
\newblock Visual genome: Connecting language and vision using crowdsourced
  dense image annotations.
\newblock {\em International Journal of Computer Vision}, 123(1):32--73, 2017.

\bibitem{liu2019learning}
Daqing Liu, Hanwang Zhang, Zheng-Jun Zha, and Wu Feng.
\newblock Learning to assemble neural module tree networks for visual
  grounding.
\newblock In {\em The IEEE International Conference on Computer Vision (ICCV)},
  2019.

\bibitem{liu2019referring}
Daqing Liu, Hanwang Zhang, Zheng-Jun Zha, and Fanglin Wang.
\newblock Referring expression grounding by marginalizing scene graph
  likelihood, 2019.

\bibitem{liu2019clevr}
Runtao Liu, Chenxi Liu, Yutong Bai, and Alan~L Yuille.
\newblock Clevr-ref+: Diagnosing visual reasoning with referring expressions.
\newblock In {\em Proceedings of the IEEE Conference on Computer Vision and
  Pattern Recognition}, pages 4185--4194, 2019.

\bibitem{luo2017comprehension}
Ruotian Luo and Gregory Shakhnarovich.
\newblock Comprehension-guided referring expressions.
\newblock In {\em Proceedings of the IEEE Conference on Computer Vision and
  Pattern Recognition}, pages 7102--7111, 2017.

\bibitem{mao2016generation}
Junhua Mao, Jonathan Huang, Alexander Toshev, Oana Camburu, Alan~L Yuille, and
  Kevin Murphy.
\newblock Generation and comprehension of unambiguous object descriptions.
\newblock In {\em Proceedings of the IEEE conference on computer vision and
  pattern recognition}, pages 11--20, 2016.

\bibitem{ren2015faster}
Shaoqing Ren, Kaiming He, Ross Girshick, and Jian Sun.
\newblock Faster r-cnn: Towards real-time object detection with region proposal
  networks.
\newblock In {\em Advances in neural information processing systems}, pages
  91--99, 2015.

\bibitem{schuster2015generating}
Sebastian Schuster, Ranjay Krishna, Angel Chang, Li Fei-Fei, and Christopher~D
  Manning.
\newblock Generating semantically precise scene graphs from textual
  descriptions for improved image retrieval.
\newblock In {\em Proceedings of the fourth workshop on vision and language},
  pages 70--80, 2015.

\bibitem{shi2019explainable}
Jiaxin Shi, Hanwang Zhang, and Juanzi Li.
\newblock Explainable and explicit visual reasoning over scene graphs.
\newblock In {\em Proceedings of the IEEE Conference on Computer Vision and
  Pattern Recognition}, pages 8376--8384, 2019.

\bibitem{yang2019cross}
Sibei Yang, Guanbin Li, and Yizhou Yu.
\newblock Cross-modal relationship inference for grounding referring
  expressions.
\newblock In {\em Proceedings of the IEEE Conference on Computer Vision and
  Pattern Recognition}, pages 4145--4154, 2019.

\bibitem{yang2019dynamic}
Sibei Yang, Guanbin Li, and Yizhou Yu.
\newblock Dynamic graph attention for referring expression comprehension.
\newblock In {\em Proceedings of the IEEE Conference on Computer Vision and
  Pattern Recognition}, 2019.

\bibitem{yang2020relationship}
Sibei Yang, Guanbin Li, and Yizhou Yu.
\newblock Relationship-embedded representation learning for grounding referring
  expressions.
\newblock {\em IEEE Transactions on Pattern Analysis and Machine Intelligence},
  2020.

\bibitem{yu2018mattnet}
Licheng Yu, Zhe Lin, Xiaohui Shen, Jimei Yang, Xin Lu, Mohit Bansal, and
  Tamara~L Berg.
\newblock Mattnet: Modular attention network for referring expression
  comprehension.
\newblock In {\em Proceedings of the IEEE Conference on Computer Vision and
  Pattern Recognition}, pages 1307--1315, 2018.

\bibitem{yu2016modeling}
Licheng Yu, Patrick Poirson, Shan Yang, Alexander~C Berg, and Tamara~L Berg.
\newblock Modeling context in referring expressions.
\newblock In {\em European Conference on Computer Vision}, pages 69--85.
  Springer, 2016.

\bibitem{zhang2018grounding}
Hanwang Zhang, Yulei Niu, and Shih-Fu Chang.
\newblock Grounding referring expressions in images by variational context.
\newblock In {\em Proceedings of the IEEE Conference on Computer Vision and
  Pattern Recognition}, pages 4158--4166, 2018.

\bibitem{zhuang2018parallel}
Bohan Zhuang, Qi Wu, Chunhua Shen, Ian Reid, and Anton van~den Hengel.
\newblock Parallel attention: A unified framework for visual object discovery
  through dialogs and queries.
\newblock In {\em Proceedings of the IEEE Conference on Computer Vision and
  Pattern Recognition}, pages 4252--4261, 2018.

\end{thebibliography}
}

\end{document}